\documentclass[10pt,twocolumn,letterpaper]{article}

\usepackage[pagenumbers]{cvpr}               %
\usepackage{amsmath}
\usepackage{amssymb}
\usepackage{graphicx}

\usepackage[usestackEOL]{stackengine}

\usepackage{verbatim}
\usepackage{xspace}

\usepackage{soul}

\newcommand{\michal}[2][]{%
    \ifthenelse{ \equal{#1}{} }
        {\textcolor{purple}{ Michal: #2}}
}\newcommand{\ljnote}[2][]{%
    \ifthenelse{ \equal{#1}{} }
        {\textcolor{purple}{ Linyi: #2}}
}

\definecolor{cvprblue}{rgb}{0.21,0.49,0.74}
\usepackage[pagebackref,breaklinks,colorlinks,allcolors=cvprblue]{hyperref}

\title{Eye2Eye: A Simple Approach for Monocular-to-Stereo Video Synthesis}
\vspace{-0.5cm}

\author{
    Michal Geyer$^{1,2}$ \hspace{5mm}
    Omer Tov$^{1}$ \hspace{5mm}
    Linyi Jin$^{1,3}$ \hspace{5mm}
    Richard Tucker$^{1}$ \\
    Inbar Mosseri$^{1}$ \hspace{5mm}
    Tali Dekel$^{1,2}$ \hspace{5mm}
    Noah Snavely$^{1}$ \\
    \vspace{3mm}
    $^{1}$Google DeepMind \hspace{10mm} $^{2}$Weizmann Institute of Science \hspace{10mm} $^{3}$University of Michigan
}

\begin{document}

\twocolumn[{%
    \renewcommand\twocolumn[2][]{#1}%
    \maketitle
    \begin{center}
        \centering \centering
        \includegraphics[width=\textwidth]{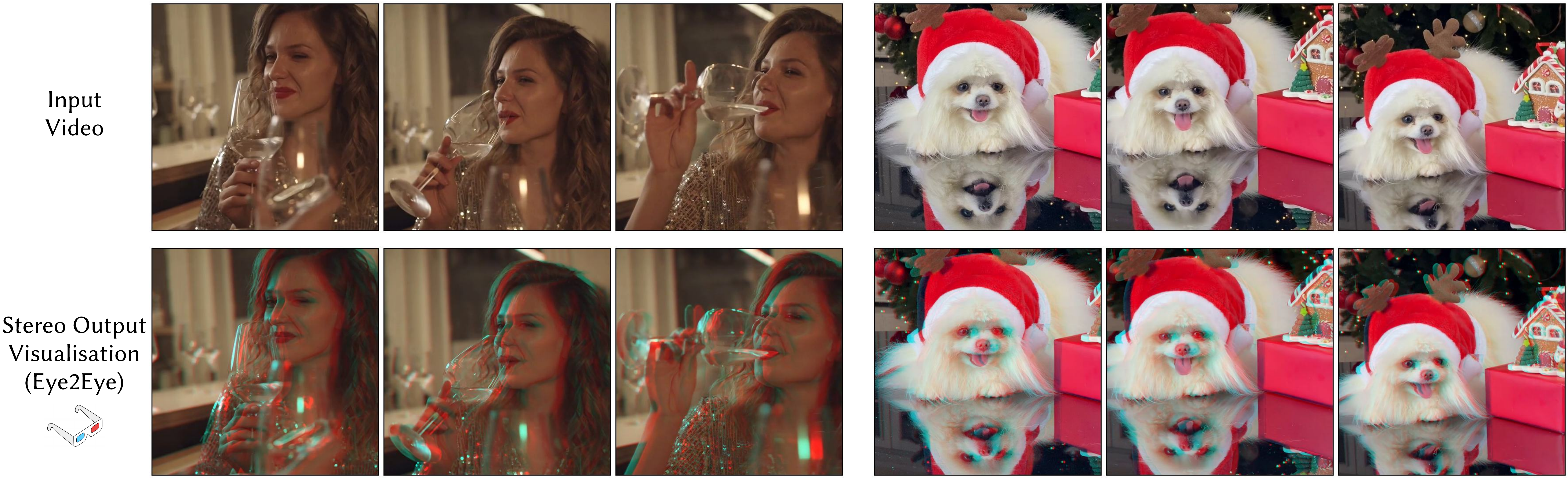}
        \vspace{-0.5cm}
        \captionof{figure}{\textbf{3D anaglyph visualization of stereo videos produced by our method.} Our framework,  \emph{Eye2Eye}, takes as input a monocular video representing a right-eye view (top), and produces a left-eye video (visualized in the anaglyph on the bottom), enabling stereoscopic viewing using 3D glasses or a VR headset. 
    Our method directly produces the new viewpoint, avoiding steps like explicit depth estimation and warping, and thus can plausibly handle challenging scenes with specular or transparent surfaces, such as the wine glass in the left example or the shiny floor in the right example, where assumptions of a single well-defined depth per pixel do not hold.}
    \label{fig:teaser}
    \end{center}
    }]

\begin{abstract}
The rising popularity of immersive visual experiences has increased interest in stereoscopic 3D video generation. 
Despite significant advances in video synthesis, creating 3D videos remains challenging due to the relative scarcity of 3D video data. 
We propose a simple approach for transforming a text-to-video generator into a video-to-stereo generator. 
Given an input video, our framework automatically produces the video frames from a shifted viewpoint, enabling a compelling 3D effect. 
Prior and concurrent approaches for this task typically 
operate in multiple phases, first estimating video disparity or depth, then warping the video accordingly to produce a second view, and finally inpainting the disoccluded regions. 
This approach inherently fails when the scene involves specular surfaces or transparent objects. 
In such cases, single-layer disparity estimation is insufficient, resulting in artifacts and incorrect pixel shifts during warping. 
Our work bypasses these restrictions by \emph{directly} synthesizing the new viewpoint, avoiding any intermediate steps. This is achieved by leveraging a pre-trained video model’s priors on geometry, object materials, optics, and semantics, without relying on external geometry models or manually disentangling geometry from the synthesis process. 
We demonstrate the advantages of our approach in complex, real-world scenarios featuring diverse object materials and compositions. See videos on \url{https://video-eye2eye.github.io/}.
\end{abstract}

\section{Introduction}

Immersive viewing hardware---such as VR headsets and 3D displays---is rapidly improving, offering users  increasingly high-quality 3D experiences. However, capturing high-quality 3D content remains challenging and often requires specialized equipment, limiting the availability of immersive media. This generates a growing demand for methods that can generate high-quality 3D content, such as stereoscopic video. Ultimately, we can envision a future where any video content can be experienced in 3D regardless of its original capture. Towards this goal, we address the problem of up-converting monocular 2D video to more immersive stereoscopic 3D video, leveraging recent advances in generative video models.

The prevalent approach to mono-to-stereo video conversion adopts a two-step process: they first estimate geometry for an input video via monocular 
depth models, 
then use this geometry to re-project pixels to a second view, inpainting dis-occluded regions to generate a video for the second eye. However, such \emph{warp-and-inpaint} approaches have an inherent restriction---they are inapplicable to scenes with reflections and complex light transport. 
Fundamentally, using a disparity map to warp an image to a new viewpoint assumes that there is a single, distinct depth at every input pixel.
For scenes that exhibit simple Lambertian reflectance, this assumption largely holds true.
However, for scenes with more complex light transport---specular reflection, partial transparency, etc.---we often cannot characterize each pixel with a single depth.
For instance, in Fig. \ref{fig:comparison}, a person is viewed through a glass window, thus the window's pixels are a mixture of the person and the objects reflected on the glass---each at completely different (virtual) depths.
To correctly handle such cases, methods based on explicit pixel warping would need to decompose the scene into multiple layers---e.g., reflected and transmitted light---warp each separately, then composite the results \cite{Sinha2012ImagebasedRF}. 
Without such handling, these methods can produce physically implausible views, for instance, where reflections appear pasted on a reflective surface, rather than at their correct virtual depth.
The effects of such artifacts
have been widely studied in cognitive science, where they have been shown to affect the way shape, material, and geometry are perceived \cite{Brain-specular-reflections, highlight-disparity, impact-of-approx,disparity-motion,role-of-specular}.

We propose to address these limitations by \emph{directly} producing the output RGB view, sidestepping the need for explicit disparity estimation or pixel warping.
We leverage recent video diffusion models for this goal, as well as the observation that while full multi-camera 3D video datasets are scarce, stereo videos captured from two-view setups are relatively abundant online. 
Such videos represent ideal training data for mono-to-stereo methods, and allow us to learn to directly produce the desired output in a way that optimizes for the actual ground truth second-eye view, no matter how complex the underlying light transport is. 
We call this method \textbf{Eye2Eye}.

Our direct approach yields superior performance over warp-and-inpaint baselines in challenging real-world scenes featuring specular or transparent surfaces and dynamic lighting conditions. We validate these findings through a user study as well as a stereo perception metric introduced in \cite{tamir2024makesgoodstereoscopicimage}. 

In summary, our contributions are:  (1) demonstrating, for the first time, mono-to-stereo video generation of specular dynamic scenes;
(2) showing how to effectivity leverage a pre-trained generative video model for this task, using curated online stereo videos; and
(3) providing quantitative and qualitative evaluations via a user study and a perceptual stereo metric that highlight the advantages of our approach over existing warp-and-inpaint methods.

\section{Related work}
\label{sec:related_work}

\paragraph{Multi-view video synthesis.}
Progress in Generative-AI has been expanded to 3D generation, with trained image and video models being repurposed for static and dynamic multi-view generation.
CAT3D \cite{cat3d} inflates an image diffusion model to take an arbitrary number of frames of a static scene as input, and to generate as output a $360^{\circ}$ set of views, from which a 3D reconstruction can be estimated using off-the-shelf methods~\cite{nerf,kerbl3Dgaussians}. A follow up work, CAT4D \cite{wu2024cat4d} expandeds the method to dynamic scenes, but as the base CAT3D model is an image model, it still lacks motion prior and fails to handle complex motion.
Other work tackles scaling video-diffusion architectures to the dynamic 4D setting 
\cite{4d-diffusion, colabdif, wu2024cat4d, zhang2024recapture}. As fully multi-view video datasets are scarce, these methods often build largely on synthetic data, static scenes and monocular videos, which limits their performance on real-world videos. 
Our two-view stereo setting allows us to use a dataset of real-world online videos from \cite{jin2024stereo4d}, enabling stereo generation of arbitrarily complex videos in terms of scene-dynamics, camera motion, and light conditions. 

\paragraph{Mono-to-Stereo Conversion.}

Early mono-to-stereo conversion methods primarily relied on motion parallax \cite{4293013}, perceptual heuristics \cite{stereobrush, Zeng2015HallucinatingSF, Kellnhofer2015SAP, 7926422}, or user interaction \cite{5928340}. 
Deep3D~\cite{Xie2016Deep3DFA} uses a CNN to predict each right video frame from the left by first estimating a soft disparity map and then compositing the output frame. 
These early approaches share a common limitation: the absence of a generative prior.

More recent mono-to-stereo synthesis methods employ a multi-stage pipeline, involving:
(1) estimating video disparity (and temporally smoothing it),
(2) using it to warp frames to the output view, and
(3) inpainting disoccluded regions.
Early works using this approach include \cite{Kopf-OneShot-2020, watson-2020-stereo-from-mono}. 
Recent methods following this pipeline build on top of generative diffusion models: StereoDiffusion \cite{Wang2024StereoDiffusionTS} for images, and SVG \cite{SVG}, StereoCrafter \cite{stereocrafter}, and SpatialDreamer \cite{lv2024spatialdreamerselfsupervisedstereovideo} for videos. 
SVG leverages a pretrained text-to-video model without any further training, by devising a specific inpainting scheme, while StereoCrafter and SpatialDreamer fine-tune an image-to-video model, modifying it (1) to be video- rather than image-conditioned, and (2) to inpaint left-right dis-occlusion regions.

Our approach offers a key advantage over those pipelines, as in many real-world scenarios a single-layered disparity estimate is insufficient to represent scene geometry. While some progress on multi-layer flow prediction has been recently made \cite{wen2024layeredflow}, correctly estimating layered \emph{video} disparity remains an overlooked challenge.
Instead, we directly leverage a pre-trained video model’s implicit, joint priors on geometry, object materials, and light, helping to alleviate this issue.

\paragraph{Novel view synthesis with reflections and specularities.}
Another possible approach for stereo synthesis is to apply a 3D video reconstruction pipeline and render stereo views from it. 
A line of work focuses on such 3D video reconstruction pipelines \cite{som2024, lee2024fastviewsynthesiscasual, tretschk2021nonrigid, wang2024gflow}.
Other work has focused on improving the ability of 3D reconstruction methods to render and reconstruct scenes with specular reflections, including: 
(1) re-parameterizing outgoing radiance as a function of the reflected view direction \cite{verbin2022refnerf, specnerf, liang2023envidr, wang2023unisdf}, 
(2) combining 3D reconstruction with inverse graphics (simultaneously estimating material properties) \cite{bi2020nrf,jin2023tensoir,srinivasan2021nerv,mai2023neural}, 
(3) directly tracing reflection rays \cite{verbin2024nerfcastingimprovedviewdependentappearance}. 
In the context of 3D \emph{video} reconstruction, a recent work incorporates physically-based rendering into a Gaussian-splatting 3D video reconstruction pipeline to handle specular reflections \cite{fan2024spectromotion}. 
In contrast, our approach leverages the implicit modeling capabilities of a large pretrained video model, eliminating the need for explicit physics-based representations. 
Furthermore, existing 4D reconstruction pipelines rely heavily on the input video to constrain the learned geometry and appearance, and often fail when the input lacks sufficient information (for example, when the camera motion is minimal, as demonstrated in Fig.~\ref{fig:comparison}). 
These limitations make 3D video reconstruction pipelines less robust for stereo generation.

\begin{figure}[htbp]
    \includegraphics[width=\columnwidth]{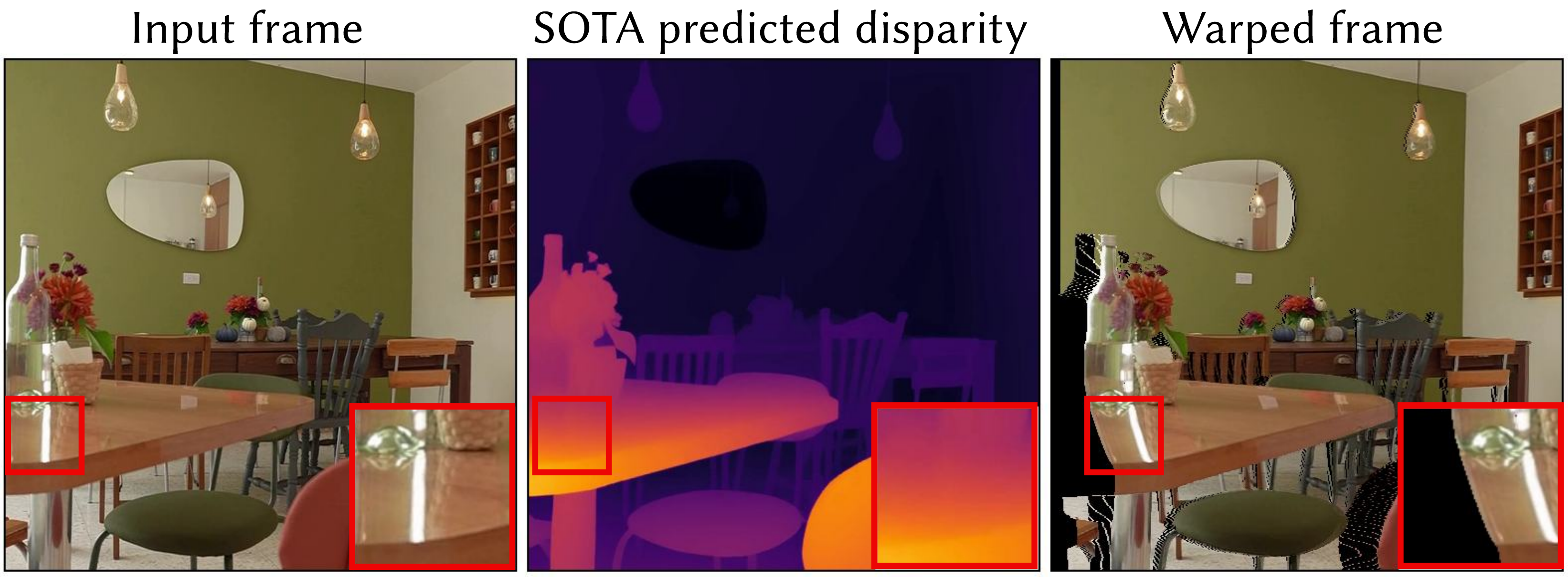}
    \caption{\textbf{Limitations of warp-and-inpaint approach for mono-to-stereo video synthesis.} Given an input video frame (left), we use a state-of-the-art disparity estimation model~\cite{hu2024-DepthCrafter} to compute its disparity (middle), and use it to warp the original frame to a new view (right). Since the predicted disparity map  captures only the 
    surface of the table, without considering the reflection of other objects off of it, the  warped frame depicts incorrect reflection (skewed diagonally instead of reflecting vertically). When viewed in VR, the reflection on the table appears ``flat'', as if it is a part of the table. This demonstrates the fundamental limitation of the common warp-and-inpaint approach for stereo view synthesis.
    }
    \label{fig:disp_estimation_problem}
\end{figure}

\section{Preliminaries}
\label{sec:prelim}
\subsection{Stereo geometry}
\label{subsec:stereo_geom}
\label{}
The geometric relationship between corresponding points in a stereo pair is governed by epipolar geometry. For a rectified stereo setup with parallel camera projection planes, a 3D point $(x, y, z)$ projects to image coordinates $(u_L, v)$ in the left view and $(u_R, v)$ in the right view, where the horizontal disparity $d = u_L - u_R$ is inversely proportional to depth: $d = \frac{fb}{z}$, where $f$ is the focal length and $b$ is the baseline distance between cameras. In the case of specular surfaces, a single depth value $z$ cannot be assigned to each pixel, since the depth of the surface itself $z_\text{surface}$ and that of the reflected object $z_\text{reflected-object}$ may differ. Thus, to correctly re-render the video from another view point, the surface and the reflected content should shift according to their depth as in the above equation: $d_\text{surface} = \frac{fb}{z_\text{surface}}$, $d_\text{reflection} = \frac{fb}{z_\text{reflected-object}}$. Fig \ref{fig:disp_estimation_problem} demonstrates how warping pixels that have reflections only with $d_\text{surface}$ distorts the rendered image.

\subsection{Diffusion models}
\label{subsec:prelim_diffusion_obj}
A diffusion model learns to reverse a noising process. Given a clean image $x_0$, the forward noising process adds Gaussian noise according to a variance schedule $\beta_t$, producing noisy samples $x_t = \sqrt{\alpha_t}x_0 + \sqrt{1-\alpha_t}\epsilon$, where $\alpha_t = \prod_{s=1}^t(1-\beta_s)$ and $\epsilon \sim \mathcal{N}(0, I)$. The simplified diffusion objective minimizes:
\begin{equation}
    \mathcal{L}_{\text{simple}} = \mathbb{E}_{t,x_0,\epsilon}\left[\|\epsilon - \epsilon_\theta(x_t, t)\|_2^2\right]
\end{equation}
When conditioned on additional inputs $c$, the model learns the conditional distribution $\epsilon_\theta(x_t, t, c)$. During inference, the model iteratively denoises random noise $x_T$ back to a clean sample. 
We build our framework on top of Lumiere \cite{bartal2024lumiere}, which is a cascaded video diffusion model. 

Cascaded diffusion models consist of two components: a base model that generates videos at low resolution, and a spatial-super-resolution (SSR) model that upsamples low-resolution outputs to a higher resolution. 
The SSR model is a conditional diffusion model that is trained to denoise high resolution videos conditioned on downsampled videos. 
At inference time, the SSR model iteratively denoises Gaussian noise into a high resolution video, conditioned on the low-resolution video produced by the base model.

\section{Method}
\label{sec:method}

\begin{figure}[htbp]
    \centering
    \includegraphics[width=\columnwidth]{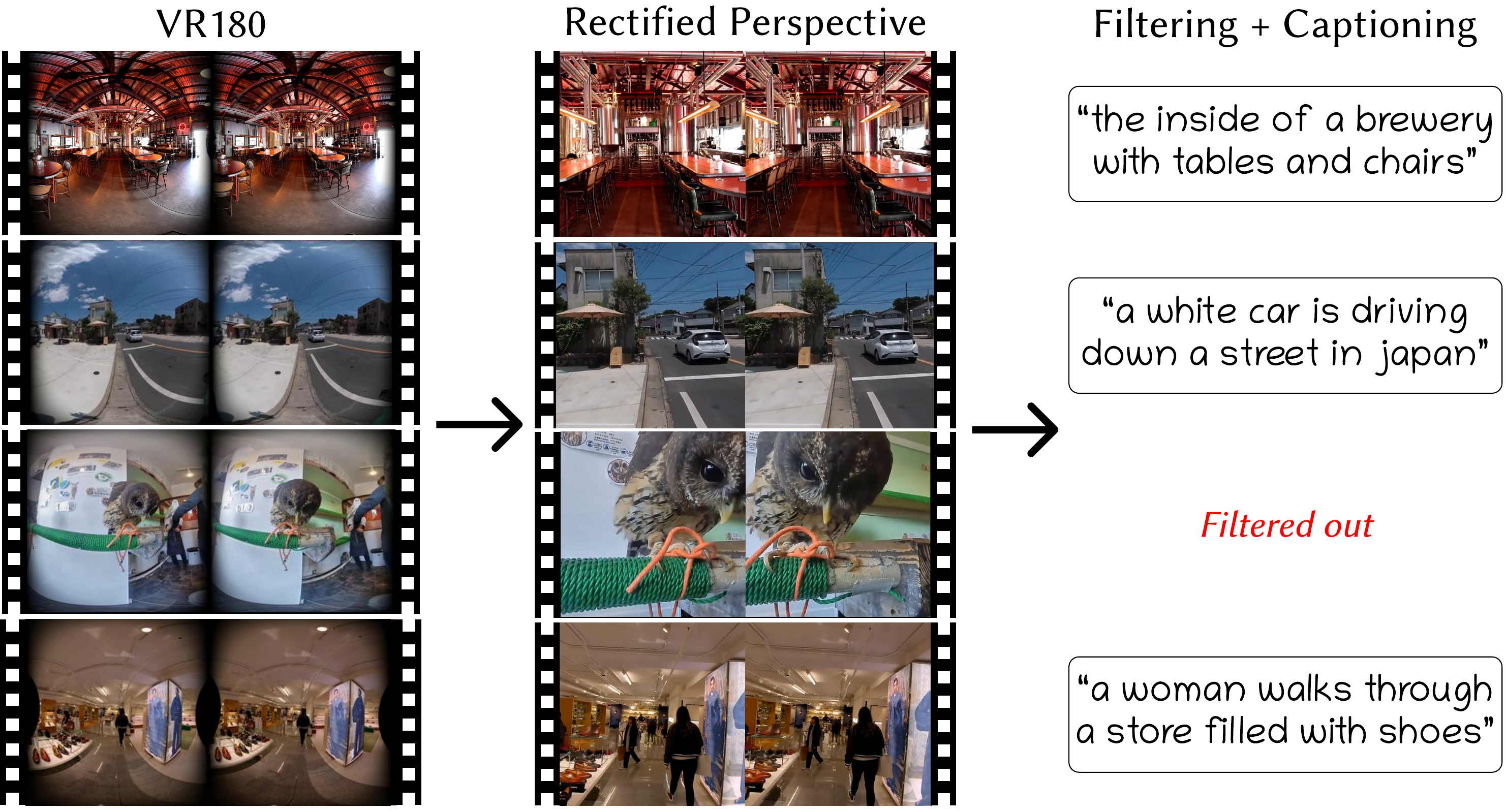}
    \caption{\textbf{Data processing pipeline.} We curate stereo VR180 footage captured with high-resolution cameras and stored in a equirectangular format. Following Stereo4D~\cite{jin2024stereo4d}, we rectify the stereo videos and map the equirectangular format to perspective videos. We filter out videos with large disparity using RAFT \cite{teed2020raft} and caption the remaining videos with BLIP2~\cite{li2023blip}.}
    \label{fig:data}
\end{figure}

Given a monocular input video $V^\mathrm{right}$, our goal is to synthesize its corresponding stereo pair by generating a left view $V^\mathrm{left}$, as if captured by a camera horizontally shifted from the original camera position by approximately human interpupillary distance (roughly 6.5cm), as per the rectified stereo geometry described in Section~\ref{sec:prelim}. 

Our task presents two key challenges: 
(1) understanding the video's geometry and light transport sufficiently well to determine how to transform the pixel content into the new view, and 
(2) synthesizing realistic content for regions that are occluded in the original view, but become visible in the new viewpoint. 
Given that generative video diffusion models have been shown to capture priors on both scene geometry and occluded content  \cite{sora,veo2024,gupta2023photorealisticvideogenerationdiffusion,kondratyuk2024videopoetlargelanguagemodel}, we propose to leverage such models to jointly address both challenges, as well as a stereo video dataset, containing left and right eye viewpoints of dynamic, in-the-wild videos. 
Specifically, we extend Lumiere \cite{bartal2024lumiere}, a cascaded text-to-video diffusion model, and construct training data from \cite{jin2024stereo4d}, to address this task. 

While we maintain Lumiere's two-stage process of low-resolution generation followed by super-resolution, we make two principal changes to adapt it to our task. 
First, our model takes a video as input, in addition to text (rather than text alone). 
Second, we find that Lumiere's super-resolution design is not suitable for stereo synthesis, leading us to develop a different approach. 
We detail these modifications in the following sections, as well as our stereo dataset collection and processing. 
We call our overall method \textbf{Eye2Eye}.

\begin{figure*}[htbp]
    \centering
    \includegraphics[width=\textwidth]{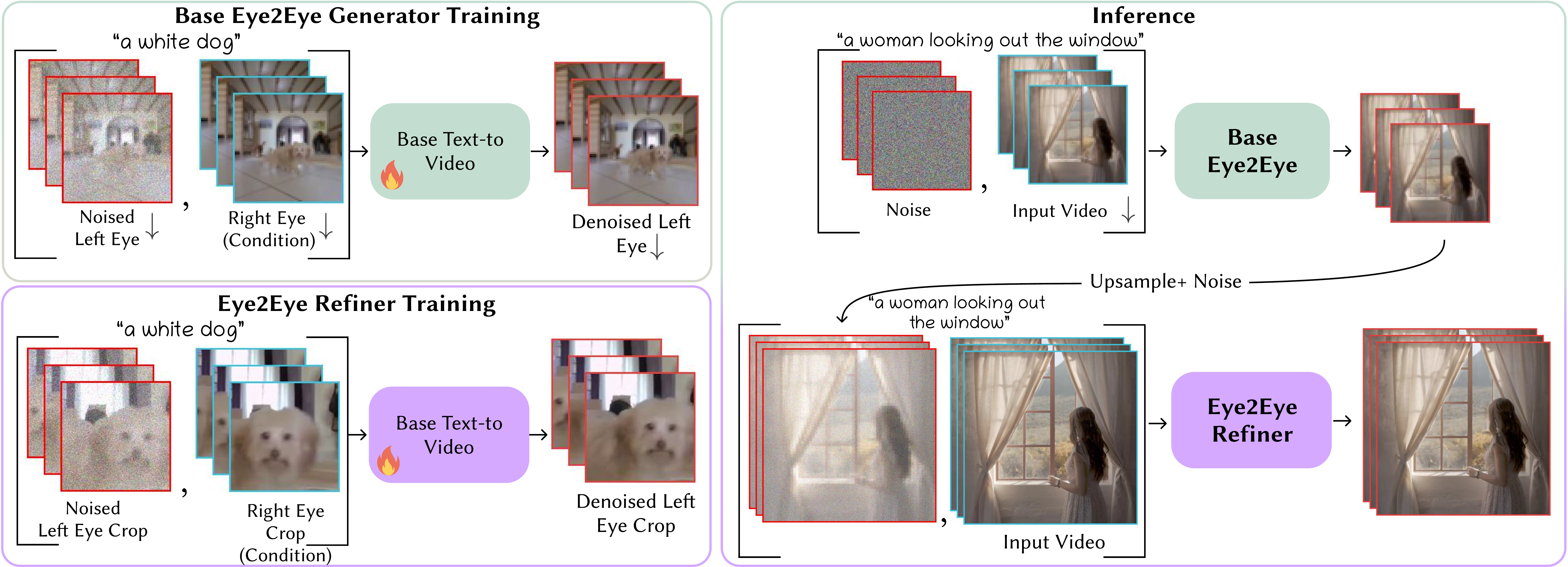}
    \caption{\textbf{Eye2Eye mono-to-stereo pipeline.} We leverage the pre-trained Lumiere cascaded text-to-video model, as well as a curated dataset of rectified stereo pairs, to perform mono-to-stereo synthesis. 
    We finetune two different copies of a base (low-resolution) pre-trained Lumiere model, in two different contexts. 
    For the first base model, we add additional input channels to condition the model on an input right eye, and train the base Eye2Eye generator on downsampled, low-resolution 128$\times$128 stereo pairs (top left). We call the resulting trained model the \textbf{base Eye2Eye generator} model.
    We train the second model to be a refinement model with the same conditioning mechanism, but instead trained on 128$\times$128 \emph{crops} from full, high-resolution images (bottom left). We call the resulting model the \textbf{Eye2Eye refiner} model.
    The base Eye2Eye model models correct pixels disparity at a low resolution, and the Eye2Eye refiner has better quality in inpainted areas or areas with large disparities. 
    At inference time, our sampling process (right) combines both models' strengths by first generating a low-resolution output from the base Eye2Eye model to establish appropriate stereo disparity for a compelling 3D effect, then noising and denoising it with the Eye2Eye refiner to achieve high visual quality.}

    \label{fig:pipeline}
\end{figure*}

\subsection{Low-resolution stereo generation}

Our first step focuses on fine-tuning the base Lumiere model $\phi(x_t, t, c)$ (where t is the diffusion timestep and c is the text conditioning) to produce left-from-right views. We do so by modifying its architecture to accept additional conditioning channels in its first input convolution layer. 
The model is trained to denoise the left view while being conditioned on the clean right view, following the standard conditional diffusion training formulation (as described in Section~\ref{sec:prelim}). 
This results in a model that produces novel left views at 128-pixel resolution (Fig.~\ref{fig:pipeline} top left). 
We call this model the \textbf{base Eye2Eye generator} and denote it by $\tilde{\phi}_\mathrm{base}$. After training, given an input down-sampled video, $V_\downarrow^{\mathrm{right}}$, and a caption $c$, $\tilde{\phi}_\mathrm{base}$ produces a low resolution left-view video: 
\begin{equation}
 \label{eq:base}
    \tilde{\phi}_\mathrm{base}(x_T, T, V_{\downarrow}^{\mathrm{right}}, c) = V^{\mathrm{left}}_\mathrm{base}
\end{equation}

\subsection{High-resolution stereo refinement} 

While the base stereo generator successfully creates left-from-right views, achieving high-resolution stereo synthesis presents additional challenges. 
Directly applying the pre-trained Lumiere super-resolution (SSR) model to $V^{\mathrm{left}}_\mathrm{base}$ would produce details inconsistent with the original input video $V^\mathrm{right}$, as the SSR denoises Gaussian noise based solely on $V^{\mathrm{left}}_\mathrm{base}$. 
We experimented with modifying the SSR model to be conditioned on the right view, but this resulted in degraded quality, which we attribute to its fully convolutional, simpler architecture compared to the base model. Therefore, we take a different route and adapt the base model for high-resolution left-from-right video synthesis.

\begin{figure}[htbp]
    \centering
    \includegraphics[width=\columnwidth]{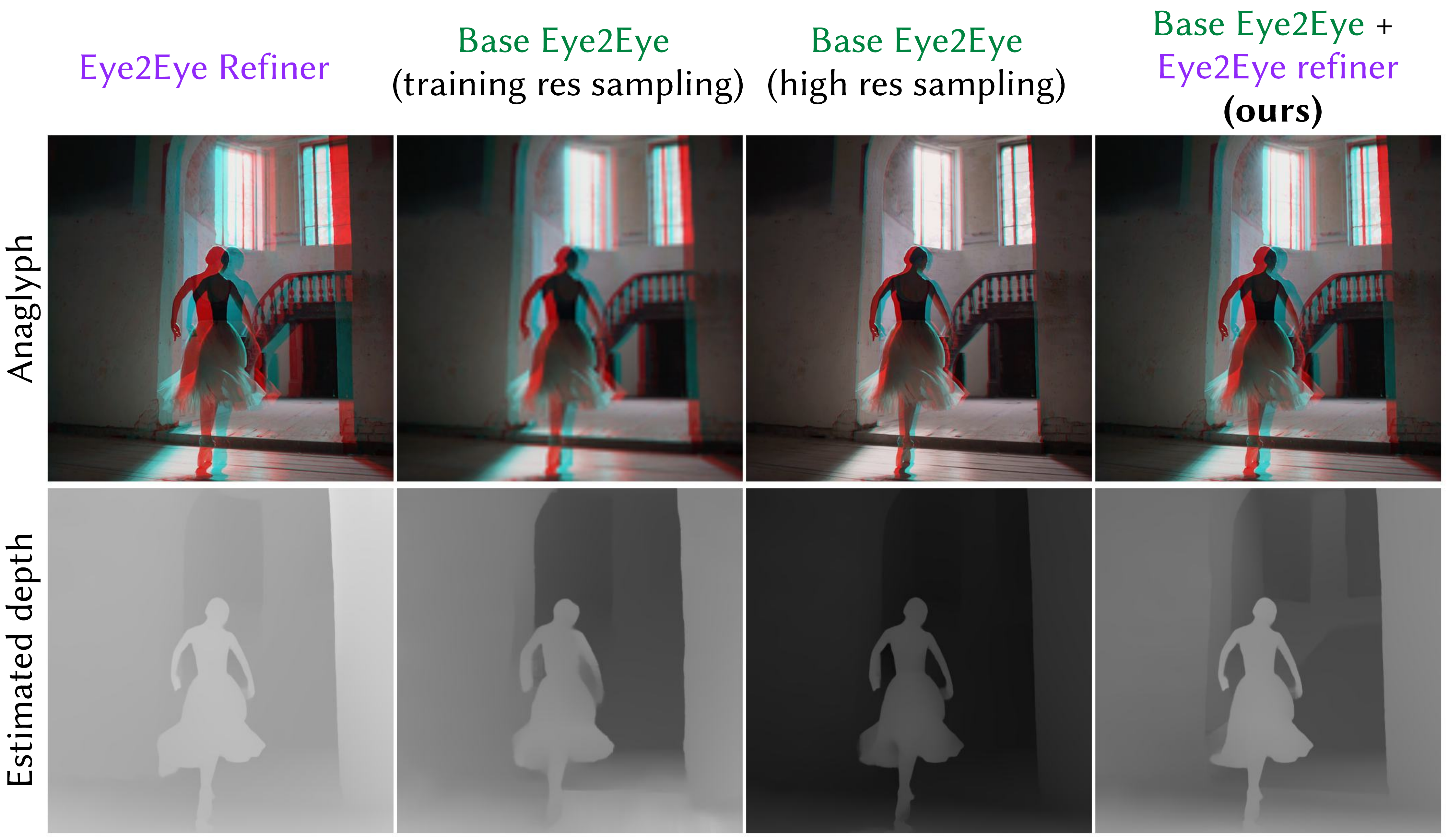}
    \caption{\textbf{Resolving training and inference gap.} We ablate the use of the two models in our pipeline, illustrating the training-inference gap of each of them, and visualize the resulting anaglyph and depth estimation (estimated using  \cite{teed2020raft}) of their outputs. 
    When sampling from the Eye2Eye-refiner model (trained on crops without any downsampling), far away content is still shifted by a large amount (column 1).
    When sampling from the base Eye2Eye generator at a higher resolution than its training resolution, the scale of the disparity and novel content in the frame reduces, weakening the 3D effect compared to sampling at the training resolution (columns 2 and 3, in column 2 the outputs were upsampled). By upsampling the outputs of the base stereo model and noising and denoising it with the Eye2Eye-refiner model, we maintain both a good depth perception from the base model and the stereo refiner's ability to generate high quality frames (column 4).}
    \label{fig:HR_disparity}
\end{figure}

Consider a pixel in a video with a resolution of 128×128 that has a disparity of $d$ pixels between the original and generated view. When generating a video at 512×512 resolution (4× higher), the disparity should scale proportionally ($4d$) pixels to maintain the same real-world depth effect. 
However, we observe that when sampling from $\tilde{\phi}_\mathrm{base}$ at different input resolutions, the pixel disparity remains at $d$ pixels rather than scaling with the resolution. 
This leads to an undesirable effect: sampling at higher resolutions effectively reduces the perceived 3D depth in the stereo pairs, as shown in Fig.~\ref{fig:HR_disparity}, columns 2 and 3. This behavior is analogous to changing the scale of the disparity itself.

To address this issue, we instead fine-tune $\phi$ on \emph{high-resolution crops} of size 128$\times$128 to learn correct-scale disparity and inpainting (Fig.~\ref{fig:pipeline} top right). 
We observe that although training on high-resolution crops indeed allows high-resolution sampling with larger pixel shifts, this approach introduces its own challenge: small crops often contain limited disparity variation and distant content. This causes the model to develop a bias toward uniformly shifting the input view (Fig.~\ref{fig:HR_disparity}, column 1).

While simply bilinearly up-sampling $V^{\mathrm{left}}_\mathrm{base}$ yields the correct disparity scale and stereo geometry, the model trained on high resolution crops yields better quality in inpainted areas or in areas where the disparity is large. To combine the strengths of both models, we use them in a two-stage inference pipeline, which exploits a fundamental property of diffusion models---early denoising steps establish global layout and structure, while later steps refine details \cite{SDEdit}. 
Specifically, we use $\tilde{\phi}_\mathrm{base}$ to produce a low-resolution layout with correctly scaled disparity, 
and use the model trained on crops as an \emph{Eye2Eye refiner model}, denoted by $\tilde{\phi}_\mathrm{refiner}$. That is, our method:
\begin{enumerate}
\item generates an initial low-resolution left view video using the base Eye2Eye generator, $V^{\mathrm{left}}_\mathrm{base}$ as in eq \ref{eq:base}, %
\item upsamples this output to the target resolution and noises the upsampled output,
\begin{equation*}
    x_t^{\mathrm{left}} = \sqrt{\alpha_t} \cdot {V^{\mathrm{left}}_\mathrm{base}}_\uparrow + \sqrt{1-\alpha_t}\cdot \epsilon
\end{equation*} where $\alpha_t$ is the diffusion noise schedule parameter as described in Sec. \ref{subsec:prelim_diffusion_obj},
\item denoises the noised upsampled-resolution video using the \emph{stereo refiner} model:
\begin{equation*}
    V^{\mathrm{left}} = \tilde{\phi}_\mathrm{refiner}(x_t^{\mathrm{left}}, t, V^{\mathrm{right}}, c)
\end{equation*}

\end{enumerate}
In other words, we perform SDEdit \cite{SDEdit} on  $V^{\mathrm{left}}_\mathrm{base}\uparrow$ with $\tilde{\phi}_\mathrm{refiner}$. This combined approach preserves correct scale disparity from the low-resolution generation while enabling high-resolution refinement of fine details and textures (Fig.~\ref{fig:HR_disparity}, rightmost column). 
The result is a pipeline that consistently balances stereo disparity with high-resolution detail, effectively bridging the gap between training and inference resolution.

\subsection{Training dataset}
\label{subsec:data}
We construct our training data from the Stereo4D~\cite{jin2024stereo4d} dataset, which contains over 100k high-resolution, rectified stereo videos capturing diverse scenes and moving objects. 
As shown in Fig.~\ref{fig:data}, this dataset provides real-world video data that naturally includes challenging cases such as reflective surfaces, which are difficult to simulate in synthetic datasets. 
Following Stereo4D, we project VR180 videos to rectified perspective 
videos of resolution 512$\times$512. We filter out examples with excessively large disparities caused by objects being too close to the camera, as these often lead to stereo window violations~\cite{zilly2010stereoscopic} and are challenging for the model to learn. 
Specifically, we compute optical flow between the left and right frames with RAFT~\cite{teed2020raft, sun2022disentangling} to estimate pixel disparities and discard videos where the disparity exceeds a specified threshold (60 pixels). 
Additionally, we use BLIP2~\cite{li2023blip} to generate captions from the middle frame of each video. During training, we sample 80 frames per clip to align with Lumiere’s input video length.

\section{Results}
\label{sec:res}

\subsection{Baselines}
The most prominent baseline for our approach is Stereo\-Crafter~\cite{stereocrafter}, which adopts a warp-and-inpaint approach and trains an inpainting model specifically for handling left-right disocclusions. 
Since StereoCrafter builds upon Video-Stable-Diffusion, a different pretrained video diffusion model than the one we use, we re-implement Stereo\-Crafter using Lumiere—the pretrained model utilized in our method, in order to ensure a fair comparison between approaches. 
We fine-tune the low-resolution Lumiere model specifically on warped views and subsequently employ the Lumiere super-resolution stage combined with a blended diffusion approach~\cite{Avrahami_2022} to preserve details from the original warped videos. We refer to this baseline 
as \emph{warp and inpaint}. See the appendix for more details.

We additionally include qualitative comparisons with Deep3D~\cite{Xie2016Deep3DFA}, a deep CNN trained for mono-to-stereo video prediction; and Dynamic Gaussian Marbles (DGM)~\cite{Stearns_2024}, a method for novel view synthesis of monocular videos.

\subsection{Evaluation data}
\begin{figure*}[htbp]
    \centering
    \includegraphics[width=\textwidth]{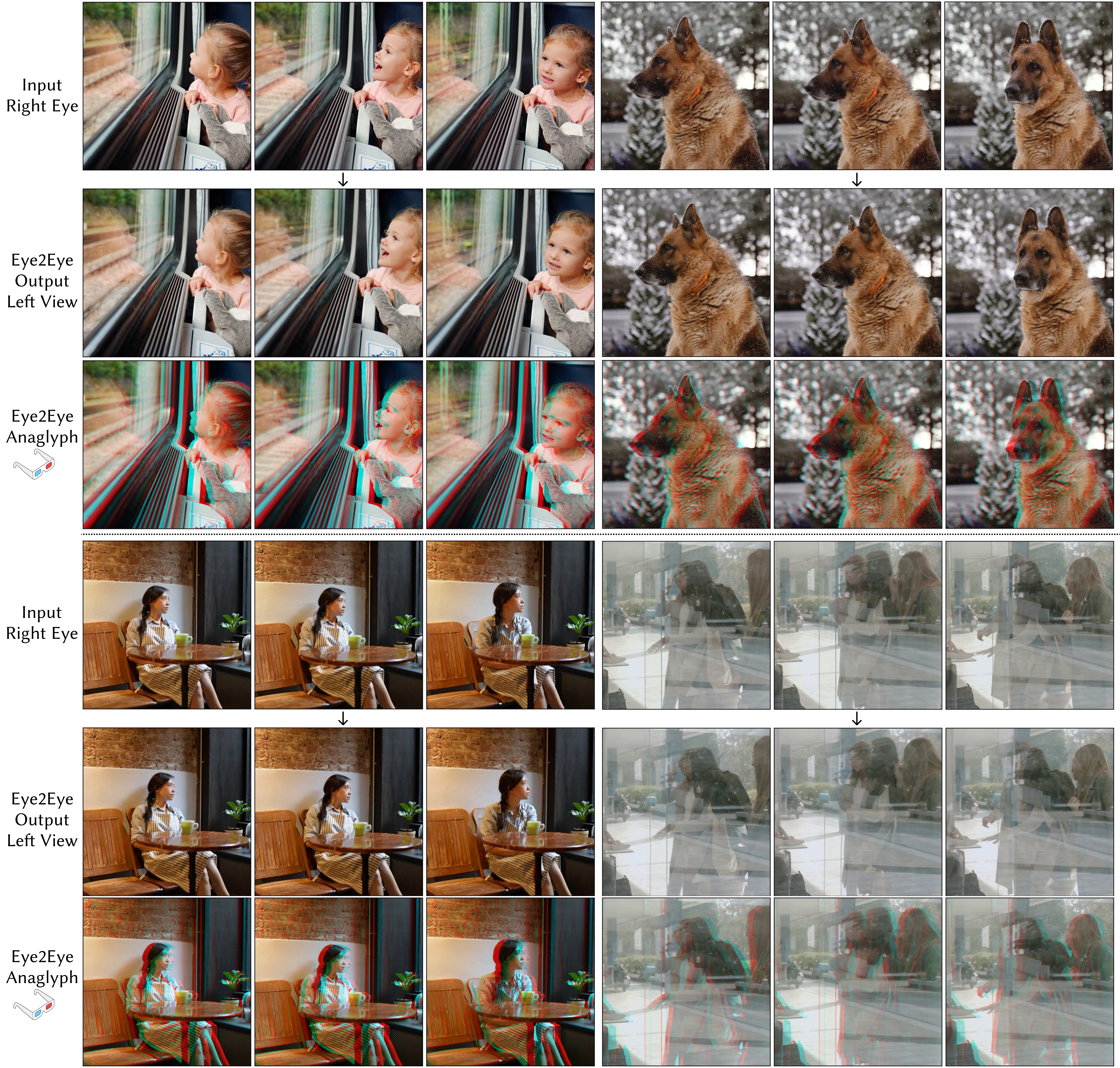}
    \caption{\textbf{Our generated stereo views.}  Our approach is particularly successful in complex scenarios involving reflective objects such as glass doors or specular tables, where traditional methods often produce distortions. See videos in our website.}
    \label{fig:results}
\end{figure*}

We assess our method on a held-out test set of 30 publicly sourced videos, encompassing diverse scenes, camera motions, and dynamic content. These videos are chosen to feature complex lighting conditions and varied materials, including specular surfaces that introduce challenges such as reflections.
Some of the videos are taken from the data provided in \cite{Xue2015ObstructionFree}, which proposed a method to decompose the different layers of reflected and refracted light. See a sample of the evaluation videos in Figure~\ref{fig:results}.

\subsection{Qualitative comparisons}
\begin{figure*}[htbp]
    \centering
    \includegraphics[width=\textwidth]{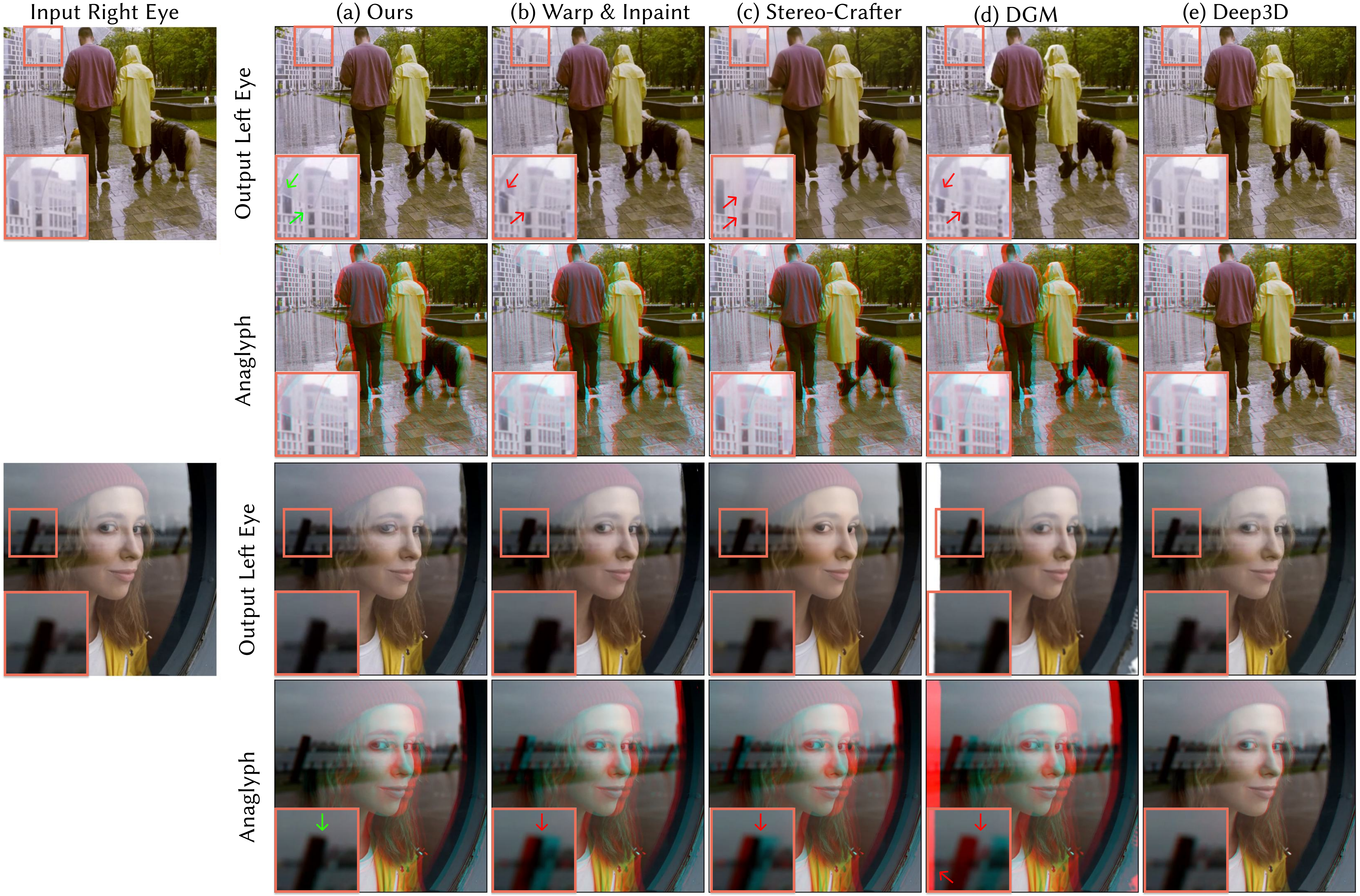}
    \captionsetup{font=small}
    \caption{\textbf{Qualitative Comparison.}
     Our method successfully generates left from right views in complex scenarios where light is both reflected on a transparent material and refracted through it. The warp-and-inpaint baseline, relying on a single-layer disparity prediction, fails in such cases. For instance, in the top example, the top of the building appears as near as the transparent umbrella overlaying it (see anaglyph), and the building is distorted (see output left eye). Our method, in contrast, successfully shifts the umbrella without shifting the building behind it. In the bottom example, the pole reflected on the glass appears as near as the woman behind the window (see anaglyph); in our result, the pole is almost not shifted, as it is far away. Dynamic Gaussian Marbles (DGM, c), a 4D reconstruction method, lacks generative capabilities. Thus, their output left eye has white regions of missing content (see top example along the borders of the people, and in the bottom example along the left edge of the frame). In addition, since DGM relies on metric depth estimation as a regularization, it often fails to correctly model the geometry in complex scenarios—producing distortions similar to those of the warp-and-inpaint baseline in the top example, and a “flat” output in the bottom example. Finally, Deep3D (d) generally fails to generate a sufficient 3D effect, as seen in the anaglyph visualizations. }
    \label{fig:comparison}
\end{figure*}

Figure~\ref{fig:comparison} shows qualitative comparisons to the baselines. 
Both the warp-and-inpaint baseline and StereoCrafter incorrectly shift scene content in areas with reflections or transparencies.
These methods struggle to handle layered structures, failing to accurately separate and shift objects with reflections or transparencies.

In the top example, depth-warping methods (column b, c) shift the upper portion of the building more than the lower part. This occurs because the top is occluded by a transparent umbrella, and the single-layer disparity model assigns it a larger disparity. 
This causes distortion and incorrect 3D effect: the top part of the building appears as close to the viewer as the umbrella when viewed with red-cyan anaglyph glasses or in VR. Similarly, in the bottom example, the reflected distant pole is shifted too much, along with the woman's head. 

In contrast, our results (column a) preserve the correct depth layering by shifting each visible layers according to its own disparity. The transparent umbrella is shifted more than the building behind it, and the reflection of the pole is shifted only slightly, while the woman's head is shifted more substantially, consistent with their relative depths.

DGM (column c) cannot inpaint missing content at occlusion boundaries since it has no generative prior, leading to holes in the video frames such as white borders near the people (top example) and along the left edge of the frame (bottom example). 
Additionally, as it uses single-layer depth estimation for geometry regularization, it also suffers from the distortions and fails to correctly model the geometry.
Deep3D (column d) fails to produce a sufficient 3D effect---the output videos are almost identical to the input ones in most cases.

\subsection{Quantitative comparisons}

\paragraph{User study.} To evaluate the benefit of our direct synthesis approach over the warp-and-inpaint approach, we conducted a user study using a Two-alternative Forced Choice (2AFC) protocol 
 \cite{kolkin2019style,park2020swapping}. Participants viewed two videos side-by-side with VR headsets: our model's output and the \emph{warp and inpaint} baseline's output. Specifically, our model's and the \emph{warp and inpaint} left view predictions were presented to the participants' left eyes, while the input right video was presented to their right eyes.

Prior to the main comparison, participants were shown a ground truth stereo pair featuring a large reflection alongside a warp-and-inpaint result that does not account for reflection. This preliminary step ensured that participants understood the task and excluded those with binocular vision dysfunctions (see the test examples in the supplemental material). 
During the main task, participants were asked to determine which video exhibited a more realistic 3D effect, including in areas with reflections or transparent surfaces. Overall, participants favored our videos $66\%$ of the time based on 239 judgments. To further statistically assess the study's results, we classified a video as favoring our method if more than half of its votes were positive (each video received between 5 and 15 votes). Out of 31 videos, 23 (about 74 $\%$) met this criterion. Under the assumption of a $50\%$ chance for a positive majority, a binomial test produced a one-sided p-value of 0.0053 (and a two-sided p-value of 0.0107), indicating that the result is statistically significant and unlikely to be due to chance.

\paragraph{iSQoE stereo perception metric} We also evaluate our results using the recently proposed stereo perception metric, iSQoE \cite{tamir2024makesgoodstereoscopicimage}, which trained a model to assess stereoscopic quality of experience (SQoE) of a stereo pair by aligning it closely with human perceptual preferences. 
The authors showed that iSQoE effectively evaluates different mono-to-stereo conversion techniques. 
iSQoE is an image (not a video) metric, and it is meaningful only when comparing the same stereo pair generated through different processing or conversion methods. Thus, to obtain per-video preferences, we average the iSQoE scores across frames and compare the mean scores between methods. Our approach achieves higher average scores on $84\%$ of the videos from our test set when compared to StereoCrafter, and $74\%$ when compared to our implemented warp-and-inpaint baseline, supporting our approach's superior performance. Interestingly, StereoCrafter performed worse than the warp-and-inpaint baseline; we attribute this to the stronger pretrained model used in our implementation (i.e., Lumiere in our warp-and-inpaint baseline vs. Stable-Video-Diffusion \cite{blattmann2023svd} in StereoCrafter).

\paragraph{}
We note that we do not report reconstruction errors with respect to the ground truth left view, since our method does not directly control the disparity scale, making pixel-wise comparisons unreliable.

\section{Discussion and Conclusions}
We presented a simple approach for video mono-to-stereo conversion, highlighting complexities that were often overlooked in prior and concurrent work. 
As video models continue to grow in size and training data, they not only produce higher-quality outputs but also appear to implicitly approximate certain aspects of our physical world; our results highlight these emerging capabilities. 
Our user study suggests handling reflections in modern VR headsets would help increase the realism of immersive experiences, encouraging VR development and research to consider these nuances. 
Nonetheless, an inherent limitation of our current approach is that we do not control the baseline between the cameras, constraining the extent of the 3D effect. Future work could explore methods for dynamically adjusting this baseline, thereby offering more flexibility in creating immersive stereo content.

\section{Acknowledgments}
We thank Shir Amir for her valuable assistance in running the iSQoE model for our evaluation.

{
    \small
    \bibliographystyle{ieeenat_fullname}
    \bibliography{references}
}

\section{Additional Details}
\subsection{Training details}
We fine-tune Lumiere on a dataset of 100K clips from Stereo4D as mentioned in Section 4.3 of the main paper. We temporally subsample the videos into 80 frames at 16 fps to match Lumiere's pre-training temporal resolution. We train the model for 120K steps with batch size 32 and learning rate $2\cdot10^{-5}$. The original clips resolution is 512$\times$512 pixels. To train the Eye2Eye base model, we additionally downsample the frames spatially to 128$\times$128 pixels. For the Eye2Eye refiner, we randomly sample crops of 128 pixels.
\subsection{Sampling hyper-parameters for our method}
\subsubsection{Base Eye2Eye sampling} 
We sample with 50 diffusion timesteps and without classifier-free guidance. We sample from this model at a resolution of 256 pixels, as we found that this resolution best mitigates visual quality and 3D effect.

\subsubsection{Eye2Eye refiner} We upsample the output of the base Eye2Eye model to 512$\times$512 pixels resolution and noise it to diffusion timestep $t=0.9$. We then denoise it with 48 diffusion timesteps and without classifier-free guidance

\section{Baselines}
\subsection{Warp-and-inpaint implementation}
\label{sec_appen:warp-inpaint}
For a fair comparison with the warp-and-inpaint approach, we implement and train this baseline using the same pretrained model as in our method.
We use the same dataset described in \ref{subsec:data} to fine tune the base Lumiere inpainting model to inpaint left-right disocclusion masks. We use \cite{teed2020raft} to estimate disparity of each pair of stereo frames, $V^\mathrm{left}, V^\mathrm{right}$ and obtain the disocclusion mask by computing left-right consistency of the disparity prediciton. At training, the model is conditioned on the right video warped according to the estimated disparity, $V_\mathrm{warped}^\mathrm{right}$, and the corresponding disocclusion mask $M$, to denoise the left frame, with the standard diffusion objective:
\begin{equation}
    \mathcal{L}_{\text{simple}} = \mathbb{E}_{t,x_0,\epsilon}\left[\|\epsilon - \epsilon_\theta(x_t, t, V_{\downarrow\mathrm{warped}}^\mathrm{right}, M, c)\|_2^2\right]
\end{equation}
Here $c$ is the text caption, $x_t = \sqrt{\alpha_t}V^\mathrm{left} +\sqrt{1-\alpha_t}\epsilon$, and $\epsilon \sim \mathcal{N}(0, I)$. Denote by $\theta(x_t, t, V_\mathrm{warped}^\mathrm{right}, M, c)$ this model after training. 
At inference time, given a video $V^\mathrm{right}$, we use SOTA monocular disparity estimation \cite{hu2024-DepthCrafter} to estimate video disparity $D^{V}$. As this estimation is scale and shift invariant, we fit a scale and shift parameter to the disparity map to align it with the disparity of our outputs (we first estimate the disparity of our outputs using \cite{teed2020raft}). We then forward-warp the frames using depth ordered softmax splatting \cite{Niklaus_CVPR_2020} and downsample the warped frames to obtain $V_{\mathrm{warped}\downarrow}^\mathrm{right}$. The inpainting mask here are the pixel locations that were not mapped onto by $D^{V}$. We open and dilate the mask to reduce temporal inconsistencies before feeding it along with the downsampled right eye video  to $\theta$ model, to obtain a low resolution inpainted video:
\begin{equation*}
    \theta(x_T, T, V_{\mathrm{warped}\downarrow}^\mathrm{right}, M, c) = V^\mathrm{inpainted}_{base}
\end{equation*}
For spatial super resolution, we use the pretrained Lumiere SSR model and take a blended diffusion approach for maintaining faithfulness to the original video. Specifically, the input to the SSR model is  the low resolution base inpainting model output $V^\mathrm{inpainted}_{base}$, and at each timestep $t$, we blend the predicted clean super-resolved output 
\begin{equation*}
    \hat{x_0^t}(x_{t}, t, V^\mathrm{inpainted}_{base})
\end{equation*}

with the high resolution warped right video
 \begin{equation*}
     V^\mathrm{right}_{\mathrm{warped}} = \mathrm{softmax\_z\_splatting}(V, D^{v})
 \end{equation*}
according to the dissocclusion mask $M$:
\begin{equation*}
    \hat{x_0^t} \leftarrow M \cdot \hat{x_0}^t(x_{t}, t, V^\mathrm{inpainted}_{base})  + (1-M) \cdot V^\mathrm{right}_{\mathrm{warped}}
\end{equation*}
This blending ensures that details in areas that appear in the input right video are preserved in the super-resolved left view.
We use a the standard lumiere sampling of 256 and 32 diffusion timesteps for the base model and the SSR model, respectively, and a classifier free guidance of 8.

\subsection{Stereo-Crafter}
We use the official Stereo-Crafter repository \url{ttps://github.com/TencentARC/StereoCrafter}. For the depth splatting stage, we scale and shift the predicted disparity in the same way described in \ref{sec_appen:warp-inpaint}.

\subsection{Deep3D}
As the original paper implementation uses a deprecated codebase, we turn to a more recent implementation found in the link: \url{https://github.com/HypoX64/Deep3D}. Their training data consists of 3D movies, which are typically processed in a different manner then our data---the zero disparity plane is usually shited to increase human comfort, making the RGB comparison difficult. We thus encourage the viewer to use anaglyph glasses for these results.

\subsection{Dynamic Gaussian marbles}
We optimize the Dynamic Gaussian Marbles using the official paper implementation \url{https://github.com/coltonstearns/dynamic-gaussian-marbles}, using their default real-world videos configuration. We observed the optimizing the representation for the full number of steps (100K) in this configuration diverges, and thus synthesize stereo views from it after 40K steps.

\end{document}